\documentclass[twoside,leqno,twocolumn]{article}

\usepackage[letterpaper]{geometry}
\usepackage{ltexpprt}
\usepackage{microtype}
\usepackage{graphicx}
\usepackage{subfigure}
\usepackage{booktabs,afterpage} 
\usepackage{xcolor}
\usepackage{soul}
\usepackage{algorithm}
\usepackage{algorithmic}
\usepackage{bm,xstring}
\usepackage{graphicx}   
\newenvironment{packeditemize}{\begin{list}{$\bullet$}{\setlength{\itemsep}{0pt}\addtolength{\labelwidth}{-5pt}\setlength{\leftmargin}{\labelwidth}\setlength{\listparindent}{\parindent}\setlength{\parsep}{0pt}\setlength{\topsep}{3pt}}}{\end{list}}
\usepackage[titletoc,title]{appendix}
\usepackage{amsmath}
\usepackage{textcomp}
\usepackage{booktabs,xspace}
\usepackage{amsfonts,verbatim}

\DeclareMathOperator{\tr}{tr}
\DeclareMathOperator{\diag}{diag}

\newcommand{\reals}{\ensuremath{\mathbb{R}}}

\newcommand{\kneteig}{KNet$_{\mathtt{EIG}}$\xspace}
\newcommand{\knetsm}{KNet$_{\mathtt{SMA}}$\xspace}
\newcommand{\kmeans}{$k$-means\xspace}

\usepackage{hyperref}

\begin{document}

\title{\Large Deep Kernel Learning for Clustering \thanks{Source code is available at \href{https://github.com/neu-spiral/kernel_net}{https://github.com/neu-spiral/kernel\_net}}}
\author{Chieh Wu \and Zulqarnain Khan \and Stratis Ioannidis \and Jennifer G. Dy \thanks{All authors are associated with Department of Electrical and Computer Engineering, Northeastern University, Boston, MA}
}

\date{}

\maketitle


\fancyfoot[R]{\scriptsize{Copyright \textcopyright\ 20XX by SIAM\\
Unauthorized reproduction of this article is prohibited}}





\begin{abstract} \small\baselineskip=9pt 
    We propose a deep learning approach for discovering kernels tailored to identifying clusters over sample data. Our neural network produces sample embeddings that are motivated by and are at least as expressive as spectral clustering.  Our training objective, based on the Hilbert Schmidt Independence Criterion, can be optimized via gradient adaptations on the Stiefel manifold, leading to significant acceleration over spectral methods relying on eigen-decompositions. Finally, our trained embedding can be directly applied to out-of-sample data. We show experimentally that our approach outperforms several state-of-the-art deep clustering methods, as well as traditional approaches such as $k$-means and spectral clustering over a broad array of real and synthetic datasets.
\end{abstract}

\section{Introduction}

   Clustering algorithms  group similar samples together based on some predefined notion of similarity. One way of representing this similarity is through kernels. However, the choice for an appropriate kernel is data-dependent; as a result, the kernel design process is frequently an art that requires intimate knowledge of the data. A common alternative is to simply use a general-purpose kernel that performs well under various conditions (e.g.,  polynomial or Gaussian kernels). 
    
    In this paper, we propose KernelNet (KNet), a methodology for learning a kernel and an induced clustering directly from the observed data. In particular, we train a \emph{deep} kernel by combining a neural network representation with a Gaussian kernel. More specifically, given a dataset $\{x_i\}_{i=1}^N$ of $N$ samples in $\reals^d$, we learn a kernel 
    $\tilde{k}(\cdot,\cdot)$  of the form:
    \begin{equation}
       \textstyle\tilde{k}(x_i,x_j) = {e^{-\frac{\|\psi_\theta(x_i) -\psi_\theta(x_j)\|_2^2}{2\sigma^2}}}/{\sqrt{d_id_j}}, 
    \end{equation}%
    where $\psi_\theta(\cdot)$ is an embedding function modeled as a neural network parametrized by $\theta$, and $\sigma$, $d_i$, $d_j$ are normalizing constants. 
    Intuitively, incorporating a neural network (NN) parameterization to a Gaussian kernel, we are able to learn a flexible deep kernel for clustering, tailored specifically to a given dataset.  
    
      \begin{figure}[!t]
    \centering
        \includegraphics[width=8cm,height=8.2cm]{{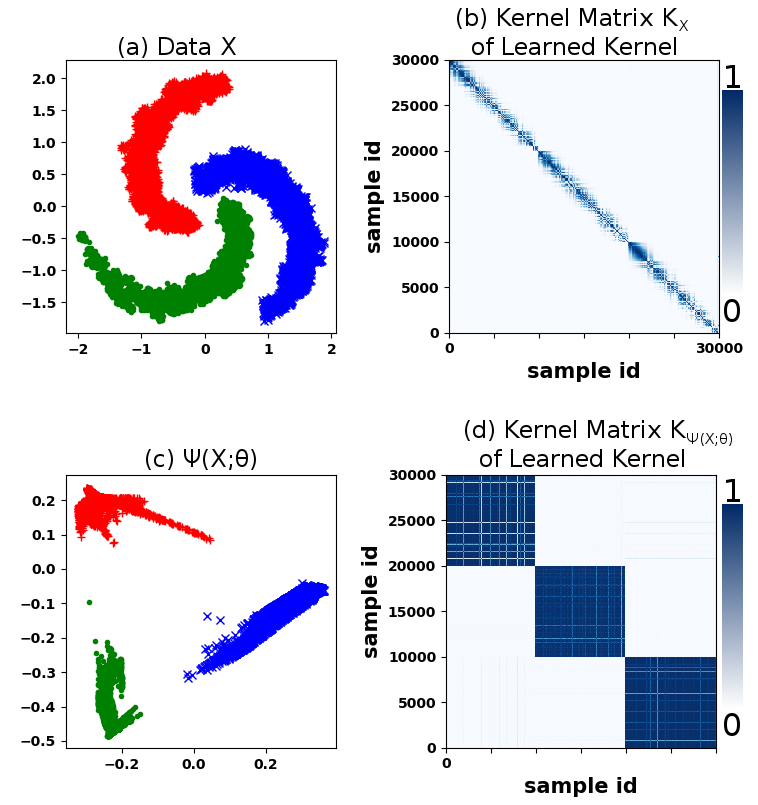}}
        \caption{Illustration of our learned embedding on the Spiral dataset. The full dataset and its kernel matrix are shown in (a) and (b), respectively. Applying a Gaussian kernel with $\sigma = 0.3$ directly to this dataset leads to a highly uninformative kernel matrix, as shown in (b). Our embedding of the entire dataset  and the resulting kernel matrix are shown in (c) and (d), respectively. Our embedding is trained only on 1\% of the samples. Yet, it generalizes to the remaining dataset, produces convex clusters, and yields an informative  kernel with a clear  block diagonal structure.}
        \label{fig:spiral_large}
    \end{figure}  
    
     We train our deep kernel with a spectral clustering objective based on the Hilbert Schmidt Independence Criterion~\cite{gretton2005measuring}. This training can be interpreted as learning a non-linear transformation $\psi$ \emph{as well as} its spectral embedding $U$ \emph{simultaneously}. Via an appropriate and intuitive initialization of our training process, we ensure that our clustering method is at least as powerful as spectral clustering. In particular, just as spectral clustering, our learned kernel and the induced clustering work exceptionally well on non-convex clusters. In practice, by training the kernel directly from the data, our proposed method significantly outperforms spectral clustering.

     The non-linear transformation $\psi$ learned directly from the dataset allows us to readily handle \emph{out-of-sample} data. Given a new unobserved sample $x_u\in \reals^d$, we can easily identify its cluster label by first computing its image $\psi_\theta(x_u)$, thus embedding it in the same space as the (already clustered) existing dataset. This is in contrast to   spectral clustering, that would require a re-execution of the algorithm from scratch on the combined dataset of $N+1$ samples. The aforementioned properties of our algorithm are illustrated in Fig.~\ref{fig:spiral_large}.
     A dataset of $N=30,000$ samples in $\reals^{2}$ with non-convex spiral clusters is shown in Fig.~\ref{fig:spiral_large}(a). Applying a Gaussian kernel with $\sigma=0.3$ directly to these samples leads to a highly uninformative kernel matrix, as shown in Fig~\ref{fig:spiral_large}(b). We train our embedding  $\psi_\theta(\cdot)$ on only 1\% of the samples, 
    and apply it to the \emph{entire dataset}; the embedded data, shown in Fig.~\ref{fig:spiral_large}(c), consists of nearly-convex,  linearly-separable clusters. More importantly, the corresponding learned kernel $\tilde{k}(\cdot,\cdot)$, yields a highly informative kernel matrix  that clearly exhibits a block diagonal structure as shown in Fig.~\ref{fig:spiral_large}(d). 
    In summary, \textbf{our major contributions are}:
     \begin{packeditemize}
    \item We propose a novel methodology of discovering a  deep kernel tailored for clustering directly from data, using an objective based on the Hilbert-Schmidt Independence Criterion.
    
    \item We propose an algorithm for training the kernel by maximizing this objective, as well as for selecting a good parameter initialization. Our algorithm, KNet, can be perceived as alternating between training the kernel and discovering its spectral embedding.

    \item We evaluate the performance of KNet with synthetic and real data compared to multiple state-of-the-art methods for deep clustering. In 5 out 6 datasets, KNet outperforms state-of-the-art by as much as 57.8\%; this discrepancy is more pronounced in datasets with non-convex clusters,  which KNet handles very well.

    \item Finally, we demonstrate that the algorithm does well in clustering out-of-sample data. This generalization capability means we can significantly accelerate KNet through subsampling: learning the embedding $\psi$ on only 1\%-35\% of the data can be used to cluster an entire dataset, leading only to a 0\%-3\% degradation of clustering performance. 
   \end{packeditemize}
 

\section{Related Work} 
\label{sec:related_work}
Several recent works propose  autoencoders specifically  designed for clustering. Song et al.~\cite{song2013auto} combine an autoencoder  with $k$-means, including an $\ell_2$-penalty w.r.t.~distance to cluster centers. They optimize this objective by alternating between stochastic gradient descent (SGD) and cluster center assignment. Ji et al.~\cite{ji2017Deep} incorporate a subspace clustering penalty to an autoencoder, and alternate between SGD and dictionary learning. Tian et al.~\cite{tian2014learning} learn a stacked autoencoder initialized via a similarity matrix. Xie et al.~\cite{xie2016unsupervised} incorporate a KL-divergence penalty  between the encoding and a soft cluster assignment, both of which are again alternately optimized; a similar approach is followed by Guo et al.~\cite{guo2017deep} and Hu et al.~\cite{hu2017learning}. 
In KNet, we significantly depart from  these methods by using an HSIC-based objective, motivated by spectral clustering. In practice, this makes KNet better tailored to learning non-convex clusters, on which the aforementioned techniques perform poorly. We demonstrate this experimentally in Section~\ref{sec:experiments}. 


Our work is closest to \emph{SpectralNet} \cite{shaham2018spectralnet} by Shaham et al. in that the authors propose a neural network approach for spectral clustering. However, they first learn a similarity matrix using a Siamese network, and then keeping this similarity \emph{fixed} they optimize a spectral clustering objective to learn the spectral embedding. In contrast, KNet 
learns \emph{both} the kernel similarity matrix and the spectral embedding \emph{jointly}, iteratively improving both.
By not having a mechanism to improve upon the previously learnt similarity matrix once a spectral embedding is learnt, as KNet does, SpectralNet can only do as well as the initially learnt similarity matrix: this is evidenced by the overall improved performance of KNet over SpectralNet (see also Sec.~\ref{sec:experiments}). 

KNet also has some relationship to methods for kernel learning. A series of papers \cite{wilson2011gaussian,wilson2016deep,wilson2016stochastic} regress deep kernels to model Gaussian processes.  Zhou et al.~\cite{zhou2004learning} learn (shallow) linear combinations of given kernels. Closest to us, Niu et al.~\cite{niu2011dimensionality} use HSIC to jointly discover a subspace in which data lives as well as its spectral embedding; the latter is used to cluster the data. This corresponds to learning a kernel over a (shallow) projection of the data to a linear subspace. KNet, therefore, generalizes the work by Niu et al.~\cite{niu2011dimensionality} to learning a deep, non-linear kernel representation (c.f.~Eq.~\eqref{eq:learnedkernel}), which improves upon spectral embeddings and is used directly to cluster the data. 

\section{Hilbert-Schmidt Independence Criterion} 
\label{sec:hsic_explained}

Proposed by Gretton et al.~\cite{gretton2005measuring}, the Hilbert Schmidt Independence Criterion (HSIC) is a statistical dependence measure between two random variables. Like Mutual Information (MI), it measures dependence by comparing the joint distribution of the two random variables with the product of their marginal distributions. However, compared to MI, HSIC is easier to compute empirically, since it does not require a direct estimation of the joint distribution.  It is used in many applications due to this advantage, including dimensionality reduction \cite{niu2011dimensionality}, feature selection \cite{song2007supervised}, and alternative clustering \cite{wu2018iterative}, to name a few.

Formally, consider a set of $N$ i.i.d.~samples $\{(x_i,y_i)\}_{i=1}^N$, where $x_i\in \reals^d$, $y_i\in \reals^c$  are drawn from a joint distribution. 
Let $X \in \mathcal{R}^{N \times d}$ and $Y  \in \mathcal{R}^{N \times c}$ be the corresponding  matrices comprising a sample in each row.
Let also $k_X: \mathbb{R}^d \times \mathbb{R}^d \rightarrow \mathbb{R}$ be any characteristic kernel, in this paper we consider Gaussian kernel $k_X(x_i,x_j) = e^{-\frac{\|x_i - x_j\|^2}{2\sigma^2}},$
and $k_Y: \mathbb{R}^c \times \mathbb{R}^c \rightarrow \mathbb{R}$ be another characteristic kernel, assumed to be the linear kernel $k_Y(y_i,y_j) = y_i^\top y_j$ here. Define $K_X,K_Y \in \mathbb{R}^{N \times N}$ to be the kernel matrices with entries $K_{X_{i,j}}=k_X(x_i,x_j)$ and $K_{Y_{i,j}} = k_Y(y_i,y_j)$, respectively, and let $\tilde{K}_X\in \reals^{N\times N }$ be the normalized Gaussian kernel matrix given by
\begin{align}
\tilde{K}_X = D^{-1/2} K_X D^{-1/2}, \label{eq:normgaus} 
\end{align}
where the degree matrix
\begin{align}D=\diag(K_X \mathbf{1}_N)\in \reals^{N\times N}\label{eq:degree_def}\end{align} is a normalizing diagonal  matrix.
Then,  the  HSIC between $X$ and $Y$ is estimated empirically via: 
\begin{align}
    \textstyle\mathbb{H}(X,Y) = \frac{1}{(N-1)^2} \tr(\tilde{K}_X H K_Y H),
    \label{eq:emprical_hsic}
\end{align}
where 
intuitively, HSIC empirically measures the dependence between samples of the two random variables. Though HSIC can be more generally defined for any arbitrary characteristic kernels,  this particular choice 
has a direct relationship with (and motivation from) spectral clustering. In particular, 
given $X$, consider the optimization: 
\begin{subequations}
            \label{eq:hsic_max_example}
    \begin{align}
         \mathop{\text{maximize}}_{U} & \hspace{0.5cm}
          \mathbb{H}(X, U)\\
        \text{subject to}  &  \hspace{0.5cm}U^\top U=I,~U\in \reals^{N\times c},
    \end{align}%
 \end{subequations}%
where $\mathbb{H}$ is given by \eqref{eq:emprical_hsic}. Then, the optimal solution $U_0\in \reals^{N\times c}$ to \eqref{eq:hsic_max_example} is precisely the spectral embedding of $X$ \cite{niu2011dimensionality}. Indeed,  $U_0$ comprises the top $c$ eigenvectors of the 
normalized similarity matrix, given by:
\begin{align}
\mathcal{L} &= H D^{-1/2} K_XD^{-1/2} H. \label{eq:laplacian}
    \end{align}
For completeness, we prove this in Appendix~\ref{HSIC_proof}.
\section{Problem Formulation} \label{sec:problem_formulation}
    We are given a dataset of samples grouped in (potentially) non-convex clusters. Our objective is to cluster samples by first embedding them into a space in which the clusters become convex. Given such an embedding, clusters can subsequently be identified via, e.g., $k$-means. We would like the embedding, modeled as a neural network, to be at least as expressive as spectral clustering: clusters separable by spectral clustering should also become separable via our embedding. In addition,  the embedding should generalize to out-of-sample data, thereby enabling us to cluster new samples outside the original dataset. \\ 
    
    \noindent\textbf{Learning the Embedding and a Deep Kernel.} Formally, we wish to identify $c$ clusters over a dataset $X \in \mathbb{R}^{N \times d}$ of $N$ samples and $d$ features. Let $\psi:\mathbb{R}^d\times \reals^m \rightarrow \mathbb{R}^{d'}$ be an embedding of a sample to $\mathbb{R}^{d'}$, modeled as a DNN parametrized by $\theta \in \reals^m$; we denote by $\psi_\theta(x)$ the image of $x\in \reals^d$ under parameters $\theta$. We also denote by $\Psi:\mathbb{R}^{N \times d}\times \mathbb{R}^m \rightarrow \mathbb{R}^{N \times d'}$ the embedding of the entire dataset induced by $\psi$, and use $\Psi_\theta(X)$ for the image of  $X$.   
    
    Let $U_0 \in \mathbb{R}^{N \times c}$ be the spectral embedding of $X$, obtained via  spectral clustering. We  can train $\psi$  to induce similar behavior as $U_0$ via the following optimization:
     \begin{align}
         \max_{\theta\in\reals^m}  
        \mathbb{H}(\Psi_\theta(X), U_0),
        \label{eq:only_HSIC_first}
    \end{align}       
    where $\mathbb{H}$ is  given by Eq.~\eqref{eq:emprical_hsic}. Since HSIC is a dependence measure, by training $\theta$ so that $\Psi_\theta(X)$ is maximally dependent on $U_0$, $\Psi$ becomes a surrogate to the spectral embedding, sharing  similar properties. 
    
    However, the surrogate $\Psi$ learned via \eqref{eq:only_HSIC_first} is restricted by $U_0$, hence it can only be as discriminative as $U_0$. To address this issue, we depart from \eqref{eq:only_HSIC_first} by jointly discovering both  $\Psi$ as well as a \emph{coupled} spectral embedding $U$. In particular, we solve the following optimization problem w.r.t. \emph{both} the embedding \emph{and} $U$:      
    \begin{subequations}
      \label{eq:psi_interpretation}
        \begin{align}
         \mathop{\text{maximize}_{\theta, U}} 
         & \hspace{0.3cm} 
        \mathbb{H}(\Psi_\theta(X), U) - \lambda \|X\!-\! f_{\theta,\theta'}(X)\|^2_2
        \label{eq:first_joint_equation}\\
        \text{subject to:}  &  \hspace{0.3cm}U^\top U=I,\label{eq:stiefel}\\ &  \hspace{0.3cm}\theta,\theta'\in \reals^m,~U\in \reals^{N\times c},   
        \end{align}%
    \end{subequations}%
where,  
\begin{align}f_{\theta,\theta'}(X)= \Psi'_{\theta'}(\Psi_\theta(X)) \label{eq:autoencoder}\end{align}
is an autoencoder, comprising $\Psi:\reals^{n\times d}\times \reals^m\to\reals^{N\times d'}$ and $\Psi':\reals^{N\times d'}\times \reals^m \to\reals^{N\times d} $ as an encoder and decoder respectively. 
This autoencoder objective is theoretically necessary to ensure that the embedding $\Psi$ is injective, as stated in Theorem 1 by Li et al. in \cite{li2017mmd}. However, as observed in \cite{li2017mmd}, our empirical experiments also suggest that this autoencoder objective is not necessary in practice: the dependence of the local minimum found by gradient descent to the starting point ensures that the trained embedding $\Psi$ is representative of the input $X$ even in the absense of this penalty (see Sec.~\ref{sec:results}).

To gain some intuition into how problem \eqref{eq:psi_interpretation} generalizes \eqref{eq:only_HSIC_first},  observe that if the embedding $\Psi$ is fixed to be the identity map (i.e., for $\Psi_\theta(X) \equiv X$) then, by  Eq.~\eqref{eq:hsic_max_example}, optimizing only for $U$ produces the spectral embedding $U_0$ .
The joint optimization of both $\Psi$ and $U$ allows us to further improve upon $U_0$, as well as on the coupled $\Psi$; we demonstrate experimentally  in Section~\ref{sec:experiments} that this significantly improves clustering quality. 

\noindent\textbf{Kernel Learning.} The optimization \eqref{eq:psi_interpretation} can also be interpreted as an instance of \emph{kernel learning}. Indeed, as discussed in the introduction, by  learning  $\psi$, we  discover in effect a normalized kernel $\tilde{k}$ of the form
\begin{align}
       \textstyle\tilde{k}(x_i,x_j) = {e^{-\frac{\|\psi_\theta(x_i) -\psi_\theta(x_j)\|_2^2}{2\sigma^2}}}/{\sqrt{d_id_j}}, \label{eq:learnedkernel} \end{align}
where $d_i,d_j$ are the corresponding diagonal elements of degree matrix $D$.


\noindent\textbf{Out-of-Sample Data.} The embedding  $\Psi$ can readily be applied to clustering out-of-sample data. In particular, having trained $\Psi$ over dataset $X$, given a new dataset $Y\in \reals^{N'\times d}$, we can  cluster this new dataset efficiently as follows. First, we use the pre-trained $\Psi$ to map every sample $y_i$ in $Y$ to its image, producing $\Psi_\theta(Y)$: this effectively embeds $Y$ to the same space as $\Psi_theta(X)$. From this point on, clusters can be recomputed efficiently via, e.g., $k$-means, or by mapping the images $\psi_\theta(y_i)$ to the closest existing cluster head. In contrast to, e.g., spectral clustering, this avoids  recomputing the joint embedding of the entire dataset $(X;Y)$ from scratch.

The ability to handle out-of-sample data can be leveraged to also accelerate training. In particular, given the original dataset $X$, computation can be sped up by training the embedding $\Psi$ by solving \eqref{eq:psi_interpretation} \emph{on a small subset of $X$}. The resulting trained $\Psi$ can be used to embed, and subsequently cluster, the entire dataset. We show in Section~\ref{sec:experiments} that this approach works very well, leading to a significant acceleration in computations without degrading clustering quality. 
  
\begin{figure}[!t]
    \centering
        \includegraphics[width=8cm,height=8cm]{{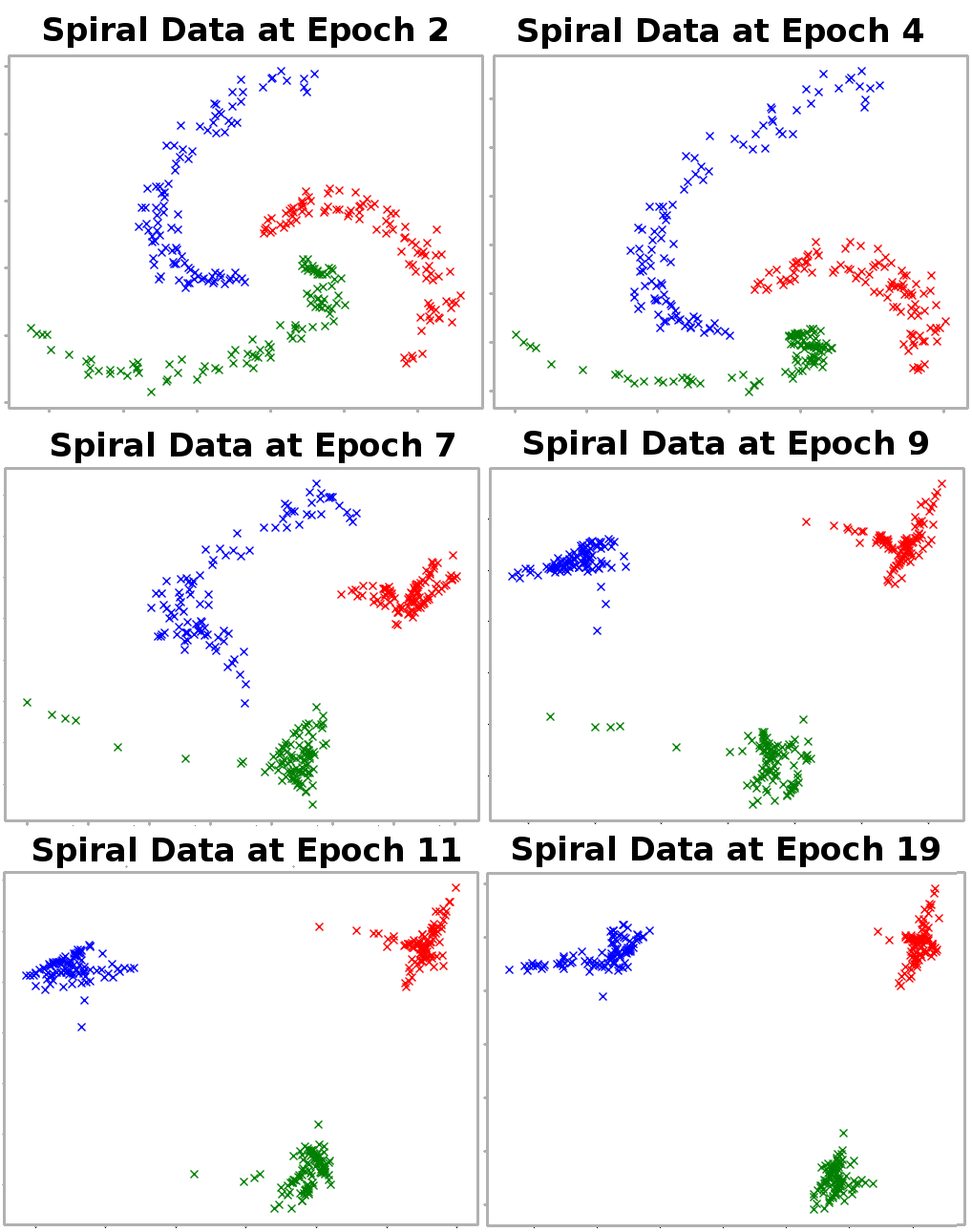}}
        \caption{Cluster images after several epochs of stochastic gradient descent over \eqref{eq:first_joint_equation}. The first term of objective \eqref{eq:first_joint_equation} pulls cluster images further apart, while making them increasingly convex. Note that this happens in a fully unsupervised fashion.}
        \label{fig:spiral_process}
    \end{figure}

\noindent\textbf{Convex Cluster Images.} The first term in objective ~\eqref{eq:first_joint_equation}  naturally encourages $\Psi$ to form convex clusters. 
To see this, as derived in Appendix~\ref{app:derivation_clumping}, ignoring the reconstruction error,  the objective \eqref{eq:first_joint_equation} becomes:
    \begin{align}
        \textstyle
        \sum_{i,j} \Gamma_{i,j} e^{-\frac{1}{2\sigma^2}
        ||\psi(\mathbf{x}_i;\theta)-\psi(\mathbf{x}_j\theta)||^2},
        \label{eq:clumping}
    \end{align}%
  where $\Gamma_{i,j}$ are the elements of matrix $\Gamma=D^{-1/2}HUU^THD^{-1/2}\in \reals^{N\times N}$.
The exponential terms in Eq.~\eqref{eq:clumping}  compel samples under which $\Gamma_{i,j}>0$  to become attracted to each other, while samples for which $\Gamma_{i,j}<0$  drift farther apart. This is illustrated in Figure \ref{fig:spiral_process}. Linearly separable, increasingly convex cluster
images arise over several iterations of solving our algorithm
at Eq.~\eqref{eq:psi_interpretation}. The algorithm, KNet, is described in the next section.

\section{KNet Algorithm}\label{sec:knet}
We solve optimization problem \eqref{eq:psi_interpretation} by iteratively adapting $\tilde{\theta}=(\theta,\theta')$ and $U$. In particular, we initialize $\Psi$ to be the identity map, and $U$ to be the spectral embedding $U_0$, and then alternate between adapting $U$ and $\tilde{\theta}.$ We optimize $\tilde{\theta}$ via stochastic gradient ascent (SGA). To optimize $U$, we adopt two approaches: one based on eigendecomposition, and one based on optimization over the Stiefel manifold.  We describe each of these steps in  detail below; a  summary can be found in Algorithm~\ref{alg:knet_algorithm}. 

\noindent\textbf{Initialization. }
The non-convexity of \eqref{eq:psi_interpretation} necessitates a principled approach for selecting good initialization points for $U$ and $\tilde{\theta}=(\theta,\theta')$. 
We initialize $U$ to $U_0$, computed via the top-$c$ eigenvectors of the 
normalized similarity matrix $\mathcal{L}$ of $X$, given by \eqref{eq:laplacian}. We initialize $\theta$ so that $\Psi$ is the identity map; 
This is accomplished by pre-training $\theta$, $\theta'$ via SGD as solutions to: 
     \begin{align} 
       \label{eq:initial_obj}
       \textstyle \mathop{\min}_{\theta,\theta'}
        \|X-\Psi_\theta(X)\|^2_2 + \|X-f_{\theta,\theta'}(X)\|^2_2.
    \end{align}   
Note that, in this construction, we use $d'=d$.
    \begin{algorithm}[!t]
    	\caption{KNet Algorithm} 
       \label{alg:knet_algorithm}
       {\small \scriptsize
    \begin{algorithmic}[1]
       \STATE {\bfseries Input:} data $X \in \mathbb{R}^{N \times d}$
       \STATE {\bfseries Output:} $\Psi$ and clustering labels
       \STATE {\bfseries Initialization:} 
       Initialize $\tilde{\theta}=(\theta, \theta')$ via  \eqref{eq:initial_obj}\\
        Initialize $U$ with $U_0$
       \REPEAT
        \STATE Update $\tilde{\theta}$ via one epoch of SGA over \eqref{eq:first_joint_equation}, holding $U$ and $D$ fixed
        \STATE  Update $D$ via Eq.~\eqref{eq:degree_def}
        \STATE \kneteig: Update $U$ via  eigendecomposition of Laplacian \eqref{eq:lappsi}, \textbf{or} \\
        \knetsm: Update $U$ via   Stiefler Manifold Ascent \eqref{eq:U_update}. 
       \UNTIL{$K_U$ has converged}
        
        \STATE Run $k$-means on $\Psi(X;\theta)$	
        \end{algorithmic}  }
      \end{algorithm}


\noindent \textbf{Updating $\mathbf{\tilde{\theta}}$.}
A simple way to update $\tilde{\theta}$ is via gradient ascent, i.e.:
\begin{align} 
       \label{eq:theta_update}
        \tilde{\theta}_k =  \tilde{\theta}_{k-1} +
        \gamma_k \nabla F(\tilde{\theta}_{k-1}, U_{k-1}),
    \end{align} 
for $k\geq 1$, where $F$ is the objective \eqref{eq:first_joint_equation}. In practice, we wish to apply stochastic gradient ascent over mini-batches; for $U$ fixed, the first term in the objective \eqref{eq:first_joint_equation} reduces to \eqref{eq:clumping}; however, the terms in the sum are coupled via
the normalizing degree matrix $D$, which depends on $\theta$ via \eqref{eq:degree_def}. This significantly increases the cost of computing mini-batch gradients. To simplify this computation, instead we hold both  $U$ \emph{and} $D$ fixed, 
and update $\tilde{\theta}$ via one epoch of SGA over \eqref{eq:first_joint_equation}. 
At the conclusion of one epoch,  we update the Gaussian kernel $K_X$ and the corresponding degree matrix $D$ via Eq.~(\ref{eq:degree_def}).  We implemented both this heuristic and regular SGA, and found that it led to a significant speedup without any observable degradation in clustering performance (see also Section~\ref{sec:experiments}). 

\noindent\textbf{Updating  $U$ via Eigendecomposition. }
Our first approach to adapting $U$ relies on the fact that, holding $\tilde{\theta}$ constant,  problem~\eqref{eq:psi_interpretation} reduces to  the form~\eqref{eq:hsic_max_example}. That is, at each  iteration, for $\tilde{\theta}$ fixed, the optimal solution $$\textstyle U^*(\tilde{\theta})=\mathop{\arg\max}_{U:U^\top U=I} F(\tilde{\theta},U)  $$ is given by the top $c$ eigenvectors of matrix 
\begin{align}
    \mathcal{L}_\theta &= H D^{-1/2} K_{\Psi_\theta(X)}D^{-1/2}H\label{eq:lappsi}.
\end{align} 
Hence, given $\tilde{\theta}$ at an iteration, we update $U$ by returning $U^*(\tilde{\theta})$. Note that when $c\ll N$, there are several algorithms for computing the top eigenvectors efficiently (see, e.g., \cite{fowlkes2004spectral, vladymyrov2016variational}).

\noindent\textbf{Updating  ${U}$ via Stiefel Manifold Ascent. } 
The manifold in $\reals^{N\times c}$ defined by constraint \eqref{eq:stiefel}, a.k.a.~the \emph{Stiefel Manifold}, is not convex; nevertheless, techniques such as those outlined in \cite{AbsMahSep2008,wen2013feasible} for optimization over this set are available in the literature. These techniques exploit the fact that descent directions that maintain feasibility can be computed efficiently. 
In particular, following \cite{wen2013feasible},
    treating $\tilde{\theta}$ as a constant, and given a feasible $U\in \reals^{N\times c}$ and the gradient of the objective $\nabla_U F(U) \in \reals^{N\times c}$ w.r.t $U$, define
    \begin{align} 
        \!A(U) \! =\! -(\nabla_U F(U) U^T \!+\! U \nabla_U F(U)^T) \in \reals^{N\times N}. 
    \end{align}         
    Using $A$ and a predefined step length $\tau$, the maximization proceeds iteratively via:
     \begin{align} 
        U_{k+1} = Q(U_k) U_k,
        \label{eq:U_update}
    \end{align}            
    where $Q$ is the so-called Cayley transform, defined as 
    \begin{align} 
        Q(U) = (I + \frac{\tau}{2}A(U))^{-1}(I - \frac{\tau}{2}A(U)). \label{eq:cayley}
    \end{align}               
    The Cayley transform satisfies several important properties \cite{wen2013feasible}. First, starting from a feasible point, it maintains feasibility over the Stiefel manifold \eqref{eq:stiefel} for all $k\geq 1$. Second, for small enough $\tau$, it is guaranteed to follow an ascent direction; combined with line-search, convergence is guaranteed to a stationary point. Finally, $Q(U_{k+1})$ given by \eqref{eq:cayley} can be computed  efficiently from $Q(U_k)$, thereby avoiding a full matrix inversion, by using the Sherman-Morrison-Woodbury identity \cite{horn1990matrix}: this  results in a $O(N^2c+c^3)$ complexity for \eqref{eq:U_update}, which is significantly faster than eigendecomposion when $c\ll N$. 
    In our second approach to updating $U$, we apply \eqref{eq:U_update} rather than eigendecomposition of $\mathcal{L}_{\theta}$ when adapting $U$ iteratively. Both approaches are summarized in line~7 of Alg.~\ref{alg:knet_algorithm}; we refer to them as \kneteig and \knetsm, respectively, in our experiments in Sec.~\ref{sec:experiments}.    


\section{Experimental Evaluation}\label{sec:experiments}
\begin{figure}[!t]
    \centering
        \includegraphics[width=8cm,height=4cm]{{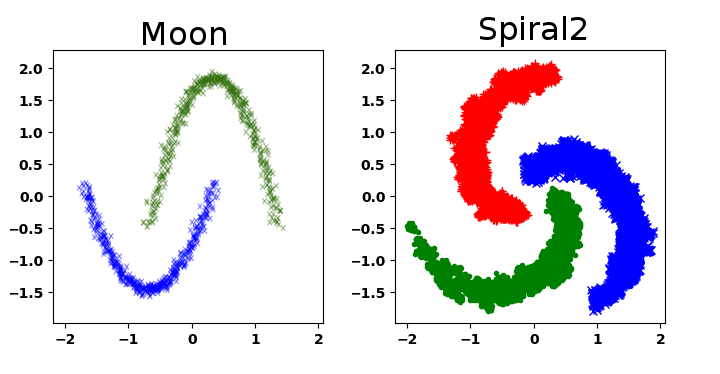}}
        \caption{Synthetic Datasets. Both dataset contain non-convex clusters. Dataset Spiral2 (depicted) contains $N=30,000$ samples, while Spiral1 contains a sub-sampled version of $N=3,000$.}
        \label{fig:moon_and_spiral}
    \end{figure}

\textbf{Datasets.}
    The datasets we use are summarized in Table~\ref{tb:dataset_info}. The first three datasets (\textbf{Moon}, \textbf{Spiral1}, \textbf{Spiral2}) are synthetic and comprise non-convex clusters;  they are shown in Figure~\ref{fig:moon_and_spiral}. Among the remaining four real-life datasets, the features of \textbf{Breast Cancer} ~\cite{breastcancer,mangasarian1990cancer} are discrete integer values between 0 to 10. The features of the \textbf{Wine} dataset ~\cite{Dua:2017} consist of a mix of real and integer values. The Reuters dataset (\textbf{RCV}) is a collection of news articles labeled by topic. We represent each  article via a \emph{tf-idf} vector using the 500 most frequent words and apply PCA  to further reduce the dimension to $d=5$. The \textbf{Face}  dataset ~\cite{bay2000uci} consists of  grey scale, $32\times30$-pixel  images of 20 faces in different orientations. We reduce the dimension to $d=20$ via PCA.
    As a final preprocessing step, we center and scale all datasets so that their mean is 0 and the standard deviation of each feature is 1.
     
    \begin{table}[!h]
    {\label{tb:dataset_info}
    \begin{center}\scriptsize
    \setlength\tabcolsep{4.0pt}
        \begin{tabular}{c|c|c|c|c|c}
        \textbf{Data}
            & \textbf{$N$}      
            & \textbf{$c$}
            & \textbf{$d$}
            & Type
            & \textbf{$\sigma$} \\
        \midrule
            \textbf{Moon}
                & 1000
            	&  2
            	& 2
            	& Geometric Shape
                & 0.1701 \\
            \textbf{Spiral1} 
                & 3000
            	& 3
            	& 2
            	& Geometric Shape
                & 0.1708 \\
            \textbf{Spiral2} 
                & 30000
            	& 3
            	& 2
            	& Geometric Shape
                & 0.1811 \\
            \textbf{Cancer} 
                & 683
            	& 2
            	& 9
            	& Medical
                & 3.3194 \\
            \textbf{Wine} 
                & 178
            	& 3
            	& 13
            	& Classification
                & 4.939 \\
            \textbf{RCV} 
                & 10000
            	& 4
            	& 5 
            	& Text 
                & 2.364 \\
            \textbf{Face} 
                & 624
            	& 20
            	&  27 
            	& Image 
                & 6.883 \\
        \midrule       
    \end{tabular}\end{center}
    }
    \caption{Dataset Summary.} 
    \end{table}    

    \begin{table*}[!t]
    {\small
    \begin{center}
    \scriptsize\setlength\tabcolsep{3.0pt}
        \begin{tabular}{c|c|c|c|c|c|c|c|c}
        \textbf{Dataset} 
            & AEC      
            & DEC     
            & IMSAT
            & SN
            & SC
            & \kmeans
            & \kneteig
            & \knetsm   \\
        \midrule
            \textbf{Moon}
            	& 56.2 $\pm$ 0.0 
            	& 42.2 $\pm$ 0.0 
            	& 51.3 $\pm$ 20.3 
             	& \textbf{100} $\pm$ 0.0
            	& 72.0 $\pm$ 0.0           	
            	& 66.1 $\pm$ 0.0
            	& \textbf{100.0 $\pm$ 0.0} 
            	& \textbf{100.0 $\pm$ 0.0} \\
            \textbf{Spiral1} 
            	& 28.3 $\pm$ 0.0 
            	& 32.0 $\pm$ 0.01 
            	& 59.6 $\pm$ 7.5 
            	& \textbf{100.0 $\pm$ 0.0}
            	& \textbf{100.0 $\pm$ 0.0}
            	& 42.0 $\pm$ 0.0
            	& \textbf{100.0 $\pm$ 0.0}
            	& \textbf{100.0 $\pm$ 0.0}\\
            \textbf{Cancer} 
            	& 79.9 $\pm$ 0.2 
            	& 79.2 $\pm$ 0.0 
            	& 74.6 $\pm$ 2.2 
             	& 82.9 $\pm$ 0.0
            	& 69.8 $\pm$ 0.0           	
            	& 73.0 $\pm$ 0.0
            	& \textbf{84.2 $\pm$ 0.4} 
            	& 82.5 $\pm$ 0.1\\
            \textbf{Wine} 
            	& 54.6 $\pm$ 0.0 
            	& 80.6 $\pm$ 0.0  
            	& 72.3 $\pm$ 11.4 
             	& 79.7 $\pm$ 0.2
            	& 88.0 $\pm$ 0.0           	
            	& 42.8 $\pm$ 0.0
            	& \textbf{91.0 $\pm$ 0.8} 
            	& 90.0 $\pm$ 0.7\\
            \textbf{RCV} 
            	& 39.3 $\pm$ 0.0
            	& 51.3 $\pm$ 0.0
            	& 39.0 $\pm$ 5.5
            	& 43.5 $\pm$ 0.2           
            	& 46 $\pm$ 0.0
            	& \textbf{56.0 $\pm$ 0}
            	& 46.3 $\pm$ 0.4
            	& 46.1 $\pm$ 0.2\\
            \textbf{Face} 
            	& 76.8 $\pm$ 0.0 
            	& 75.8 $\pm$ 1.6 
            	& 83.8 $\pm$ 3.5 
             	& 75.6 $\pm$ 0.1
            	& 66.0 $\pm$ 0.4           	
            	& 91.8 $\pm$ 0.0
            	& \textbf{93.0 $\pm$ 0.3} 
            	& 92.6 $\pm$ 0.5 \\
        \midrule       
    \end{tabular}\end{center}
    }
    \caption{The \textbf{clustering results} measured by NMI as percentages are shown above where the best mean results are highlighted in bold text. Besides the RCV dataset, KNet generally outperforms competing methods by a significant margin. The improvement is especially large with the Moon and Spiral1 dataset due to KNet's ability to identify non-convex clusters.} 
    \label{tb:full_dataset}
    \end{table*}

    \begin{table*}[!t]
    {
        \small
      \begin{center}
       \scriptsize\setlength\tabcolsep{4.0pt}
      \begin{tabular}{c|c|c|c|c|c|c|c|c|c|c|c|c}
        \toprule
        &
        \multicolumn{2}{|c}{AEC} &
        \multicolumn{2}{|c}{DEC} &
        \multicolumn{2}{|c}{IMSAT} &
        \multicolumn{1}{|c}{SN} &     
        \multicolumn{1}{|c}{SC} &
        \multicolumn{1}{|c}{\kmeans} &
        \multicolumn{3}{|c}{KN} \\
        \midrule
        \textbf{Dataset} 
            & Prep
            & RT
            & Prep
            & RT
            & Prep
            & RT
            & RT          
            & RT
            & RT      
            & Prep
            & RT$_{\texttt{EIG}}$
            & RT$_{\texttt{SMA}}$\\
        \midrule
            \textbf{Moon} 
             	& 18.23
            	& 28.04
             	& 331.80
            	& 3.52
             	& 1.20
            	& 34.34 
            	& 324.00
            	& 0.34
            	& 0.04
            	& 129.00          
            	& 28.90
            	& 18.40\\
            \textbf{Spiral1} 
              	& 2574.00 
            	& 7920.00 
             	& 343.80 
            	& 121.20 
             	& 53.06
            	& 309.00
            	& 444.00
            	& 4.20
            	& 0.12
            	& 300.00 
            	& 42.00
            	& 19.00\\
             \textbf{Cancer}
              	& 99.00
            	& 280.20
             	& 342.00
            	& 38.73
             	& 1.10
            	& 144.00
            	& 234.00
            	& 0.18
            	& 0.03
            	& 150.00
            	& 19.30
            	& 10.30\\
            \textbf{Wine}
              	& 33.81
            	& 41.69
             	& 345.00
            	& 38.02
             	& 1.10
            	& 33.32 
            	& 330.00
            	& 0.03
            	& 0.06
            	& 462.00
            	& 7.20
            	& 3.40\\
            \textbf{RCV}
              	& 141.60
            	& 2536.80
             	& 381.00
            	& 784.20
             	& 10.39
            	& 73.20
            	& 438.00
            	& 83.40
            	& 0.35
            	& 1080.00
            	& 1830.00
            	& 1116.00\\
            \textbf{Face} 
              	& 27.50
            	& 189.00
             	& 344.40
            	& 254.10
             	& 1.20
            	& 121.80
            	& 170.90
            	& 0.26
            	& 0.15
            	& 1320.00
            	& 20.90
            	& 3.30\\
        \bottomrule
      \end{tabular}\end{center}
      }
      \caption{The \textbf{preprocessing (Prep) and runtime (RT)} for all benchmark algorithms are displayed in seconds. The table demonstrates that KNet's speed is comparable to competing methods.} 
    \label{tb:fullset_time}
    \end{table*}

\noindent\textbf{Clustering Algorithms. }
    We evaluate 8 algorithms, including our two versions of KNet described in Alg.~\ref{alg:knet_algorithm}.  For  existing algorithms, we use architecture designs (e.g., depth, width) as recommended by the respective authors during training. 
    We provide  details for each algorithm below.
    
    \noindent \textbf{\kmeans: } We use the CUDA implementation\footnote{\href{https://github.com/src-d/kmcuda}{https://github.com/src-d/kmcuda}} by \cite{ding2015yinyang}.\\
    \textbf{SC: }  We use the python scikit  implementation of the classic spectral clustering algorithm by \cite{ng2002spectral}.\\
     \textbf{AEC: } Proposed by \cite{song2013auto}, this algorithm incorporates a $k$-means objective in an autoencoder. As suggested by the authors, we use 3 hidden layers of width 1000, 250, and 50, respectively, with an output layer dimension of 10.\footnote{\href{https://github.com/developfeng/DeepClustering}{https://github.com/developfeng/DeepClustering}} \\
     \textbf{DEC: } Proposed by \cite{xie2016unsupervised}, this algorithm couples an autoencoder with a soft cluster assignment via a KL-divergence penalty. As recommended, we use 3 hidden layers of width 500, 500, and 2000 with an output layer dimension of 10.\footnote{\href{https://github.com/XifengGuo/DEC-keras}{https://github.com/XifengGuo/DEC-keras}}\\
     \textbf{IMSAT: } Proposed by \cite{hu2017learning}, this algorithm trains a network adversarially by generating augmented datasets. It uses 2 hidden layers of width 1200 each, the output layer size equals the number of clusters $c$.\footnote{\href{https://github.com/weihua916/imsat}{https://github.com/weihua916/imsat}}\\
     \textbf{SN: } Proposed by \cite{shaham2018spectralnet}, SN uses an objective motivated by spectral clustering to map to a target similarity matrix.\footnote{\href{https://github.com/KlugerLab/SpectralNet}{https://github.com/KlugerLab/SpectralNet}} 
     \\
     \textbf{\kneteig } and \textbf{\knetsm: } These are the two versions of KNet, as described in Alg.~\ref{alg:knet_algorithm}, in which  $U$ is updated via eigendecomposition and Stiefel Manifold Ascent, respectively. For both versions,  the encoder and decoder have 3 layers. For Cancer, Wine, RCV, and Face dataset, we set the width of all hidden layers to $d$. For Moon and Spiral1, we set the width of all hidden layers to 20. We set the Gaussian kernel $\sigma$ to be  median of the pairwise Euclidean distance between samples in each dataset (see Table~\ref{tb:dataset_info}).\footnote{\href{https://github.com/neu-spiral/kernel_net}{https://github.com/neu-spiral/kernel\_net}}
     
     
     
\noindent\textbf{Evaluation Metrics.}
    We evaluate the clustering quality of each algorithm by comparing the clustering assignment generated to the  ground truth assignment via the Normalized Mutual Information (NMI). NMI is a similarity metric lying in $[0,1]$, with 0 denoting no similarity and 1 as an identical match between the assignments. Originally recommended by \cite{strehl2002cluster}, this statistic has been widely used for clustering quality validation \cite{niu2011dimensionality,dang2010generation,wu2018iterative,ross2013nonparametric}. We provide a formal definition in Appendix \ref{app:NMI} in the supplement. 
    
    For each algorithm, we also  measure the execution time, separating it into preprocessing time (Prep) and runtime (RT); in doing so, we separately evaluate the cost of, e.g., parameter initialization from training.

\noindent\textbf{Experimental Setup. } We execute all algorithms over a machine with 16 dual-core CPUs (Intel 
 Xeon$^\text{\textregistered}$  E5-2630 v3 @ 2.40GHz) with 32 GB of RAM with a NVIDIA 1b80 GPU. For methods we can parallelize  either over the GPU or over the 16 CPUs (IMSAT,SN,\kmeans,KNet), we ran both executions and recorded the fastest time. The code provided for DEC could only be parallelized over the GPU, while methods AEC and SC could only be executed over the CPUs.  For each dataset in Table~\ref{tb:full_dataset}, we run all algorithms on the full dataset 10 times and report the mean and standard deviation of the NMI of their resulting clustering performance against the ground truth. As SGA is randomized, we repeat experiments 10 times and report NMI averages and standard deviations.

For algorithms that can be executed out-of-sample (AEC, DEC, IMSAT, SN, KNET), we repeat the above experiment by training the embedding only on a random subset of the dataset. Subsequently, we apply the trained embedding to the entire dataset and cluster it via \kmeans. For comparison, we also find cluster centers via \kmeans on the subset and new samples to the nearest cluster center. For each dataset, we set the size of the subset (reported in Table~\ref{tb:out-of-sample_results})  so that the spectrum of the resulting subset is close, in the $\ell_\infty$ sense, to the spectrum of $X$.

    \begin{table}[!t]
    {\small
    \begin{center}
    \scriptsize\setlength\tabcolsep{1.4pt}
        \begin{tabular}{c|c|c|c|c|c|c|c|c}
         \textbf{Dataset} 
            & Data \% 
            & AEC      
            & DEC     
            & IMSAT
            & SN
            & \kmeans
            & \kneteig
            & \knetsm \\
        \midrule
            \textbf{Moon}
            	&  25\% 
            	& 51.1 
            	& 45.6 
            	& 45.5 
            	& \textbf{100.0}
            	& 21.5
            	& \textbf{100.0}
            	& \textbf{100.0}\\
            \textbf{Spiral1} 
            	& 10.0\% 
            	& 32.2
            	& 49.5
            	& 48.7
            	& \textbf{100.0}
            	& 56.7
            	& \textbf{100.0}
            	& \textbf{100.0}\\
            \textbf{Cancer} 
            	& 30.0\% 
            	& 76.4 
            	& 76.9
            	& 74.9 
            	& 82.2
            	& \text{Fails}
            	& \textbf{84.0}
            	& 83.6\\
            \textbf{Wine} 
            	& 75.0\% 
            	& 49.1 
            	& 81.5 
            	& 69.4 
            	& 77.2
            	& 25.0
            	& \textbf{91.1}
            	& 89.3\\
            \textbf{RCV} 
            	& 6.0\% 
            	& 26.8 
            	& 43.2 
            	& 35.2
            	& 41.3
            	& \textbf{52.5}
            	& 45.2
            	& 43.1\\
            \textbf{Face} 
            	& 35.0\% 
            	& 52.5 
            	& 67.1 
            	& 77.8 
            	& 75.5
            	& 87.2
            	& \textbf{92.7}
            	& 91.1\\
        \midrule       
    \end{tabular}\end{center}
    }
    \caption{
    The \textbf{out-of-sample clustering} result measured by NMI as percentages are shown above where the best mean results are highlighted in bold text. All algorithms are trained on a subset of data. We report the results of the total dataset clustered out-of-sample via each algorithm. 
    }    
    \label{tb:out-of-sample_results}
    \vspace{-1em}   

    \end{table}

    \begin{table*}[!t]
{      \begin{center}
        \scriptsize\setlength\tabcolsep{3.0pt}
        \small
      \begin{tabular}{c|c|c|c|c|c|c|c|c|c|c|c}
        \toprule
        &
        \multicolumn{2}{|c}{AEC} &
        \multicolumn{2}{|c}{DEC} &
        \multicolumn{2}{|c}{IMSAT} &
        \multicolumn{1}{|c}{SN} &
        \multicolumn{1}{|c}{\kmeans} &
        \multicolumn{3}{|c}{KN} \\
        \midrule
        \textbf{Data} 
            & Prep
            & RT
            & Prep
            & RT
            & Prep
            & RT
            & RT          
            & RT      
            & Prep
            & RT$_{\texttt{EIG}}$
            & RT$_{\texttt{SMA}}$\\
        \midrule
            \textbf{Moon} 
             	& 15.01
            	& 22.30
             	& 201.00 
             	& 2.95
            	& 1.10
            	& 27.63
            	& 140.10 
            	& 0.03s
            	& 48.00
            	& 2.30
            	& 1.10 \\
            \textbf{Spiral1} 
              	& 192.00 
            	& 498.60
             	& 250.20 
             	& 99.00
            	& 47.05
            	& 229.80
            	& 223.30
            	& 0.07
            	& 82.00
            	& 3.10
            	& 1.40  \\
             \textbf{Cancer}
              	& 55.30
            	& 75.00
             	& 306.00
             	& 30.01
             	& 1.65
            	& 100.20
            	& 38.90
            	& 0.03
            	& 16.00
            	& 4.00
            	& 1.80  \\
            \textbf{Wine}
              	& 25.30
            	& 35.31
             	& 279.00 
             	& 32.31
            	& 2.40
            	& 25.60
            	& 20.40
            	& 0.03
            	& 74.00
            	& 1.30
            	& 0.09    \\
            \textbf{RCV}
              	& 63.00
            	& 316.20
             	& 339.00
             	& 672.60
            	& 3.21
            	& 56.00
            	& 40.67
            	& 0.03
            	& 72.00
            	& 5.50
            	& 2.10 \\
            \textbf{Face} 
              	& 15.10
            	& 171.60 
             	& 315.00 
             	& 233.40 
             	& 1.10
            	& 93.60 
            	& 35.09
            	& 0.10
            	& 76.80
            	& 2.20
            	& 0.96 \\
        \bottomrule
      \end{tabular}\end{center}
      }
      \caption{ 
      The \textbf{out-of-sample preprocessing (Prep) and runtime (RT)} for all benchmark algorithms are displayed in seconds. The table demonstrates that KNet's speed is comparable to competing methods. 
      }
      \label{tb:out_of_sample_time}     
    \end{table*}

\subsection{Results}\label{sec:results}
    \begin{table}[!h]
    {\scriptsize
    \begin{center}
    \setlength\tabcolsep{4.0pt}
        \begin{tabular}{c|c|c|c|c|c|c}
        & \multicolumn{3}{c}{\kneteig}
        & \multicolumn{3}{|c}{\knetsm}\\
        \midrule
            \textbf{$\lambda$}      
            & \textbf{HSIC}
            & \textbf{AE error}
            & \textbf{NMI}
            & \textbf{HSIC}
            & \textbf{AE error}
            & \textbf{NMI}\\           
        \midrule
            	$10^{0}$
            	& 98.11	
            	& 21.58	
            	& 0.90
                & 92.01	
                & 8.98	
                & 0.89	\\
            	$10^{-1}$
            	& 98.80	
            	& 72.95	
            	& 0.88
            	& 94.21	
            	& 72.62	
            	& 0.87\\
            	$10^{-2}$
            	& 112.32	
            	& 101.45	
            	& 0.87
            	& 101.11	
            	& 88.39	
            	& 0.89\\
           	0.005
            	& 109.01	
            	& 105.37	
            	& 0.91
            	& 108.22	
            	& 107.22	
            	& 0.90\\
            	$10^{-4}$
            	& 113.18	
            	& 124.56	
            	& 0.91
            	& 110.18	
            	& 126.63	
            	& 0.91\\
            0
                & 110.89	
                & 127.23	
                & 0.92
                & 109.36	
                & 127.11	
                & 0.91\\
        \midrule  
        \end{tabular}
    \end{center}
    }
    \caption{ HSIC, AE reconstruction error, and NMI at convergence of KNet on the Wine dataset, as a function of $\lambda$.} 
    \label{tb:lambda}
    \end{table}          
     \textbf{Selecting $\mathbf{\lambda}$.} 
     As clustering is unsupervised, we cannot rely on ground truth labels to identify the best hyperparameter $\lambda$. We, therefore, need  an unsupervised method for selecting this value. 
     We find that, in practice, just like \cite{li2017mmd}, selecting $\lambda=0$ works quite well.  Because the problem is not convex, local optima reached by KNet depend highly  on the initialization. Initializing (a) $\Psi$ to approximate the identity map via \eqref{eq:initial_obj}, and (b) $U$ to be the spectral embedding $U_0$ indeed leads to a local maximum that is highly dependent on the input $X$, eschewing the need for the reconstruction error in the objective \eqref{eq:psi_interpretation}. 
     
     Our choie of $\lambda=0$ is further grounded in experiments shown in Table \ref{tb:lambda} which shows results for the Wine dataset. We found that at smaller values of $\lambda$ result in improved performance both in terms of HSIC and NMI; tables for additional datasets can be found in Appendix~\ref{app:lambda} in the supplement. Alongwith the HSIC we also provide AE reconstruction error at convergence.  Beyond the good performance of $\lambda=0$, the table suggests that an alternative  unsupervised method is to select $\lambda$ so that the ratio of the two terms at convergence is close to one. Of course, this comes at the cost of parameter exploration; in the remainder of this section, we report the performance of KNet with $\lambda=0$. \\

\noindent\textbf{Comparison Against State-of-the-Art. } Table~\ref{tb:full_dataset} shows the NMI performance of different algorithms over different datasets. With the exception of RCV dataset, we see that KNet outperforms every algorithm in Table~\ref{tb:full_dataset}. AEC, DEC, and IMSAT perform especially poorly when encountering non-convex clusters as shown in the first two rows of the table.  Spectral clustering (SC), and SN that is also based on a spectral-clustering motivated objective, perform equally well as KNet on discovering non-convex clusters. Nevertheless, KNet outperforms them for real datasets: e.g., for the Face dataset, \kneteig surpasses SN by 28\%. Note that, for the RCV dataset, \kmeans  outperformed all methods, though overall performance is quite poor; a reason for this may be the poor quality of features extracted via TFIDFs and PCA.

 KNet's ability to handle non-convex clusters is evident in the improvement over $k$-means to KNet for the first two datasets. The kernel matrix $K_X$ shown in Fig.~\ref{fig:spiral_large}(b) illustrates why \kmeans performs poorly on this dataset. 
 In contrast, the increasingly convex cluster images learned by KNet, as shown in Fig.~\ref{fig:spiral_large}(c), lead to much better separability. This is consistently observed for both the Moon and Spiral1 dataset, for which KNet achieves NMI; we elaborate on this further in Appendix~\ref{app:moon_and_spiral}. demonstrate KNet's ability to generate convex representations even when the initial representation is non-convex. 

We also note that KNet consistently outperforms spectral clustering. This is worth noting because, as discussed in Sec.~\ref{sec:problem_formulation}, KNet's initialization of both  $\Psi$ and $U$ are tied to the spectral embedding. 
Table~\ref{tb:full_dataset} indicates that alternatively learning both the kernel and the corresponding spectral embedding indeed leads to improved clustering performance.

Table~\ref{tb:fullset_time} shows the time performance of each algorithm. In terms of total time, KNet is faster than AEC and DEC. We also observe that SN is faster than most algorithms in terms of run time. However, SN does require extensive hyperparameter tuning to reach the reported NMI performance (see App.~\ref{app:learning_rate}). 
We note that a significant percentage of the total time for KNet is spent in the preprocessing step, with \knetsm being faster than \kneteig. This is due to the initialization of $\Psi$, i.e., training the corresponding autoencoder.  Improving this initialization process could dramatically speed up the total runtime. Alternatively, as we discuss in the next section, using only a small subset to train the embedding and clustering out-of-sample can also significantly accelerate the total runtime, without a considerable NMI degradation. \\

\noindent\textbf{Out-of-Sample Clustering. }
    We report out-of-sample clustering NMI performance in Table~\ref{tb:out-of-sample_results}; note that SC cannot be executed out-of-sample. Each algorithm is trained using only a subset of samples, whose size is indicated on the table's first column.  Once trained, we report in the table the clustering quality of applying each algorithm to the full set without retraining. 
    We observe that, with the exception of RCV, KNet clearly outperforms all benchmark algorithms in terms of clustering quality. This implies that KNet is capable of generalizing the results by using as litle as 6\% of the data. 
    
     By comparing Table~\ref{tb:full_dataset} against \ref{tb:out-of-sample_results}, we see that AEC, DEC, and IMSAT suffer a significant drop in performance, while KNet suffers only a maximum degradation of 3\%. Therefore, training on a small subset of the data not only yields high-quality results, the results are almost as good as training on the full set itself. Table~\ref{tb:out_of_sample_time}, reporting corresponding times, indicates that this can also lead to a significant acceleration, especially of the preprocessing step.  Together, these two observations indicate that KNet can indeed be applied to clustering of large non-convex datasets by training the embedding on only a small subset of the provided samples.

\section{Conclusions}
   \label{sec:conclusion}
   KNet performs unsupervised kernel discovery using only a subset of the data. By discovering a kernel that optimizes the Spectral Clustering objective, it simultaneously discovers an approximation of its embedding through a DNN. Furthermore, experimental results have confirmed that KNet can be trained using only a subset.

\clearpage
\newpage

\begin{appendices}
\section{Relating HSIC to Spectral Clustering }
    \begin{proof}
    Using Eq.~(\ref{eq:emprical_hsic}) to compute HSIC emprically, Eq~(\ref{eq:hsic_max_example}) can be rewritten as
    \begin{align} \label{eq:HSIC_def}
        Y^* = \arg\max_{Y} Tr(D^{-1/2}K_{X}D^{-1/2}HK_YH), 
    \end{align}      
    where $D^{-1/2}K_{X}D^{-1/2}$ and $K_Y$ are the kernel matrices computed from $X$ and $Y$. As shown by \cite{niu2011dimensionality}, if we let $K_Y$ be a linear kernel such that $K_Y=YY^T$, add the constraint such that $Y^TY=I$ and rotate the trace terms we get
    \begin{subequations}
    \begin{align} 
       \label{eq:kernel_net_objective_2}
       [Y^*] =& \arg\max_{Y} Tr( Y^THD^{-1/2}K_{X} D^{-1/2}H Y)\\
       \text{s.t : } & Y^TY=I.
    \end{align}
    \end{subequations}          
    By setting the Laplacian as $\mathcal{L}=HD^{-1/2}K_XD^{-1/2}H$, the formulation becomes identical to Spectral Clustering as
    \begin{subequations}
    \begin{align} 
       \label{eq:kernel_net_objective_3}
       [Y^*] =& \arg\max_{Y} Tr( Y^T \mathcal{L} Y)\\
       \text{s.t : } & Y^TY=I.
    \end{align}
    \end{subequations}             
    \end{proof}
\label{HSIC_proof}
\end{appendices}

\begin{appendices}
\section{Effect of $\lambda$ on HSIC, AE reconstruction error, and NMI}
    \begin{table}[!h]
    {\small
    \begin{center}
    \setlength\tabcolsep{3.0pt}
        \begin{tabular}{c|c|c|c|c|c|c}
        & \multicolumn{3}{c}{\kneteig}
        & \multicolumn{3}{|c}{\knetsm}\\
        \midrule
            \textbf{$\lambda$}      
            & \textbf{HSIC}
            & \textbf{AE error}
            & \textbf{NMI}
            & \textbf{HSIC}
            & \textbf{AE error}
            & \textbf{NMI}\\           
        \midrule
            	$10^{0}$
            	& 2.989	
            	&0.001	
            	& 0.446 
            	& 2.989	
            	& 0	
            	& 0.421 \\
            	$10^{-1}$
            	& 2.989	
            	& 0.0001	
            	& 0.412 
            	& 2.989	
            	& 0	
            	& 0.414 \\
            	$10^{-2}$
            	& 2.989	
            	& 0.0001	
            	& 0.421 
                & 2.989	
            	& 0	
            	& 0.418 \\           	
            	$10^{-3}$           
            	& 2.989	                
            	& 0.001	
            	& 0.434 
                & 2.989	
            	& 0	
            	& 0.419 \\           	
            	$10^{-4}$
            	& 2.988	
            	& 0.001	
            	& 0.421 
            	& 2.988	
            	& 0	
            	& 0.418 \\
            	$10^{-5}$
            	& 2.965	
            	& 0.099	
            	& 0.557 
                & 2.979	
            	& 0.033	
            	& 0.558 \\           	
            	$10^{-6}$
            	& 1.982	
            	& 0.652	
            	& 0.644 
                & 2.33	
            	& 0.583	
            	& 0.56 \\           	
            	$10^{-7}$
            	& 2.124	
            	& 0.669	
            	& 0.969 
                & 2.037	
            	& 0.66	
            	& 0.824 \\           	
            	$10^{-8}$
            	& 2.249	
            	& 0.69	
            	& 1 
                & 2.368	
            	& 0.67	
            	& 1 \\           	
            	0
            	& 2.278	
            	& 0.715	
            	& 1 
                & 2.121	
            	& 0.718	
            	& 1 \\           	
        \midrule  
        \end{tabular}
    \end{center}
    }
    \caption{ HSIC, AE reconstruction error, and NMI at convergence of KNet on the SPIRAL1 dataset, as a function of $\lambda$. } 
    \label{tb:lambda_in_appendix}
    \end{table}    
\label{app:lambda}
\end{appendices}

\begin{appendices}
\section{Derivation for Eq.~(\ref{eq:clumping}) }
\label{app:derivation_clumping}
By ignoring the reconstruction error, the loss from Eq.~(\ref{eq:psi_interpretation}) becomes
\begin{equation}
\mathbb{H}(\Psi_\theta(X), U).
\end{equation}
From \cite{gretton2005measuring}, this expression can be expanded into
\begin{equation}
Tr(K_{\Psi_{\theta}(X)} D^{-1/2} HK_UH D^{-1/2}).
\end{equation}
Since $D^{-1/2} HK_UH D^{-1/2}$ is a constant matrix, we let it equal to $\Gamma$, and the objective becomes
\begin{equation}
Tr(\Gamma K_{\Psi_{\theta}(X)}).
\end{equation}
The trace term can be converted into a matrix sum as
\begin{equation}
\sum_{i,j}
\Gamma_{i,j} 
K_{\Psi_{\theta}(X), i, j}.
\end{equation}
By replacing the kernel with the Gaussian kernel, Eq.~(\ref{eq:clumping}) emerges as 
    \begin{align}
        \textstyle
        \sum_{i,j} \Gamma_{i,j} e^{-\frac{1}{2\sigma^2}
        ||\psi(\mathbf{x}_i;\theta)-\psi(\mathbf{x}_j\theta)||^2},
        \label{eq:clumping_app}
    \end{align}%
\end{appendices}

\begin{appendices}
\section{Normalized Mutual Information}\label{app:NMI}
Consider two clustering assignments assigning labels in $\{1,\ldots,c\}$ to samples in dataset $\Omega=\{1,\ldots, N\}$. We represent these two assignments through two partitions of $\Omega$, namely $\mathcal{C}=\{C_\ell\}_{\ell=1}^{c}$, $\mathcal{C}'=\{C_\ell'\}_{\ell=1}^{c}$, where $$C_\ell,C_\ell'\subseteq\Omega, \ell=1,\ldots,c$$
are the sets of samples receiving label $\ell$ under each assignment.
Define the empirical distribution of labels  to be:
\begin{align}\mathbf{P}(\ell,\ell') = \frac{|C_\ell \cap C_\ell' |}{N}\quad \text{for all}~(\ell,\ell')\in \{1,\ldots,c\}^2.\label{eq:joint} \end{align}
The NMI is then given by the ratio 
$$\frac{I(\mathcal{C},\mathcal{C'})}{\sqrt{H(\mathcal{C})H(\mathcal{C}')}}$$
where $I(\mathcal{C},\mathcal{C'})$ is the mutual information and $H(\mathcal{C}),H(\mathcal{C'})$ are the entropies of the marginals of the joint distribution \eqref{eq:joint}.

\end{appendices}

\begin{appendices}
\section{Moon and Spiral dataset}
    \label{app:moon_and_spiral}
    In this appendix, we illustrate that for the Moon and the Spiral datasets, we are able to learn (a) convex images for the clusters and (b) kernels that produces block diagonal structures. The kernel matrix is constructed with a Gaussian kernel, therefore the values are between 0 to 1. The kernel matrices shown in the figures below use white as 0 and dark blue as 1; all values in between are shown as a gradient between the two colors.
    
\begin{figure}[!h]
    \centering
        \includegraphics[width=8cm,height=8cm]{{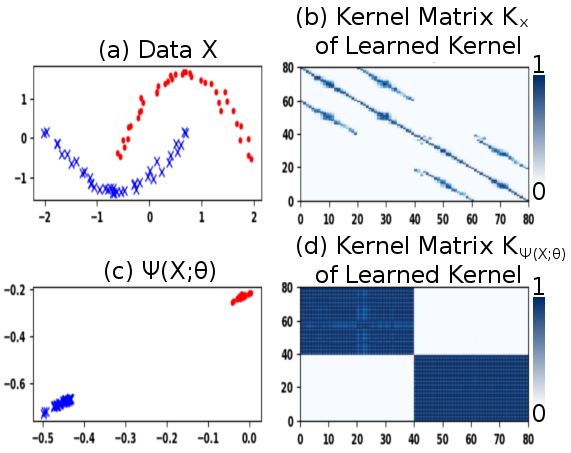}}
        \caption{This figure plots out the effect of training $\Psi$ on the Moon dataset. The original data and its kernel matrix is shown in (a) and (b) respectively. The embedding of the data with $\Psi$ and its kernel matrix is shown in (c) and (d) respectively. This figure demonstrates KNet's ability to maps non-convex clustering into convex representation.}
        \label{fig:moon_no_noise}
    \end{figure}

\begin{figure}[!t]
        \centering 
        \includegraphics[width=8cm,height=8cm]{{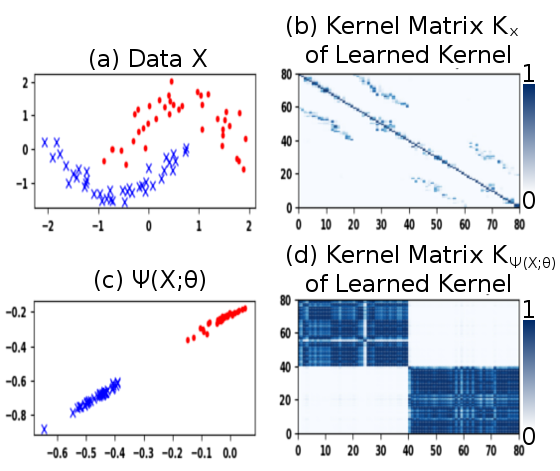}}
        \caption{This figure plots out the effect of applying a trained $\Psi$ on a distorted Moon dataset. The distorted data and its kernel matrix is shown in (a) and (b) respectively. The embedding of the distorted data with $\Psi$ and its kernel matrix is shown in (c) and (d) respectively. This figure demonstrates KNet's robustness in mapping non-convex clustering into convex representation under Gaussian distortion.}
        \label{fig:moon_noise}
    \end{figure}

    In Figure~\ref{fig:moon_no_noise}, the Moon dataset $X$ is plotted in Fig.~\ref{fig:moon_no_noise}(a) and its kernel block structure in Fig.~\ref{fig:moon_no_noise}(b). After training $\Psi$, the image of $\Psi$ is shown in Fig.~\ref{fig:moon_no_noise}(c) along with its block diagonal structure in Fig.~\ref{fig:moon_no_noise}(d). Using the same $\Psi$ trained on $X$, we distorted $X$ with Gaussian noise and plot it in Fig.~\ref{fig:moon_noise}(a) along with its kernel matrix in Fig.~\ref{fig:moon_noise}(b). We then pass the distorted $X$ into $\Psi$ and plot the resulting image in Fig.~\ref{fig:moon_noise}(c) along with its kernel matrix in Fig.~\ref{fig:moon_noise}(d). From this example, we demonstrate KNet's ability to embed data into convex clusters even under Gaussian noise.

\begin{figure}[!t]
    \centering
        \includegraphics[width=8cm,height=8.0cm]{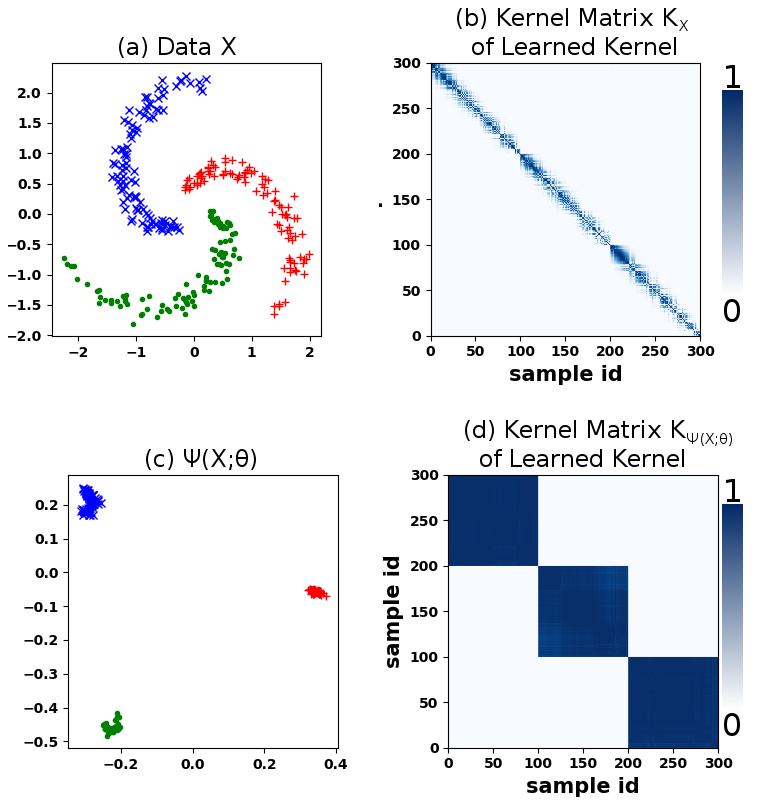}
        \caption{This figure plots out the effect of training and applying $\Psi$ on a small subset of the Spiral dataset. The subset and its kernel matrix is shown in (a) and (b) respectively. The embedding of the subset and its kernel matrix is shown in (c) and (d) respectively. This figure demonstrates KNet's ability to map non-convex cluster into convex representation with only a small subset of the data.}
        \label{fig:app:spiral_small}
    \end{figure}

\begin{figure}[t!]
     \centering
         \includegraphics[width=8cm,height=8.0cm]{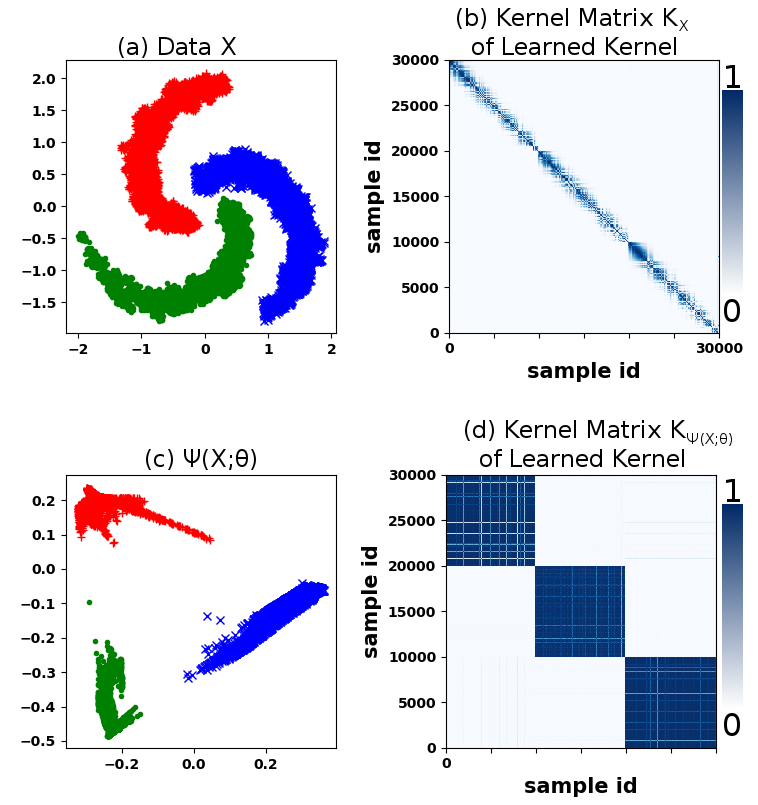}
         \caption{This figure plots out the effect of applying $\Psi$ trained on a small subset to the full Spiral dataset. The full dataset and its kernel matrix is shown in (a) and (b) respectively. The embedding of the full dataset and its kernel matrix is shown in (c) and (d) respectively. This figure demonstrates KNet's ability to generate convex representation on a large scale while training only on a small subset of the data. In this particular case, 1\%.}
         \label{fig:app:spiral_large}
     \end{figure}

    In Figure~\ref{fig:app:spiral_small}, a subset of 300 samples of the Spiral dataset is plotted in Fig.~\ref{fig:app:spiral_small}(a) and its kernel block structure in Fig.~\ref{fig:app:spiral_small}(b). After training $\Psi$, the image of $\Psi$ is shown in Fig.~\ref{fig:app:spiral_small}(c) along with its block diagonal structure in Fig.~\ref{fig:app:spiral_small}(d). The full dataset is shown in (a) of Figure~\ref{fig:app:spiral_large} along with its kernel matrix in Fig.~\ref{fig:app:spiral_large}(b). Using the same $\Psi$ trained from Fig.~\ref{fig:app:spiral_small}(a), we pass the full dataset into $\Psi$ and plot the resulting image in Fig.~\ref{fig:app:spiral_small}(b) along with its kernel matrix in Fig.~\ref{fig:app:spiral_small}(d). From this example, we demonstrate KNet's ability to generalize convex embedding using only 1\% of the data.
    
    \label{app:moon_noise}
\end{appendices}

\begin{appendices}
\section{Algorithm Hyperparameter Details}
    \begin{table}[!h]
    {\small
    \begin{center}
    \setlength\tabcolsep{4.0pt}
        \begin{tabular}{c|c|c|c}
            \textbf{Algorithm}      
            & \textbf{Learning Rate}
            & \textbf{Batch size}
            & \textbf{Optimizer}\\
        \midrule
                AEC
                & -
                & 100
                & SGD\\
                DEC
                & 0.01
                & 256
                & SGDSolver\\
                IMSAT
                & 0.002
                & 250
                & Adam\\                
                SN
                & 0.001
                & 128
                & RMSProp\\        
                KNet
                & 0.001
                & 5
                & Adam\\            
        \midrule  
        \end{tabular}
    \end{center}
    }
    \caption{ Here we include the typically used learning rates and batch sizes, and the optimizer type for each algorithm. These were set as recommended by the respective papers, except in the case of AEC which is silent on what learning rate needs to be set, the available implementation sets the learning rate with a line search. We use the above mentioned settings generally and only change them if the batch size is too big for a dataset or we notice the preset learning rate not leading to convergence.} 
    \label{tb:learning_rate}
    \end{table}    
\label{app:learning_rate}

The hyperparameters for each network are set as outlined in the respective papers. In the case of SN, the hyper-parameters included the number of neighbors for calculation, the number of neighbors to use for graph Laplacian affinity matrix, the number of neighbors to use to calculate the scale of the Gaussian graph Laplacian, and the threshold for calculating the closest neighbors in the Siamese network. They were set by doing a grid search over values ranging from 2 to 10 and by using the loss over $10\%$ of the training data.
\end{appendices}



\begin{thebibliography}{99}

\bibitem{gretton2005measuring}
A.~Gretton, O.~Bousquet, A.~Smola, and B.~Sch{\"o}lkopf,
  {\em Measuring statistical dependence with Hilbert-Schmidt norms}, International Conference on Algorithmic Learning Theory (2005),
  pp.~63--77.
  
\bibitem{song2013auto}
C.~Song, F.~Liu, Y.~Huang, L.~Wang, and T.~Tan,
  {\em Auto-encoder based data clustering},
  Iberoamerican Congress on Pattern Recognition (2013),
  pp.~117--124.

\bibitem{ji2017Deep}
P.~Ji, T.~Zhang, H.~Li, M.~Salzmann and I.~Reid,
  {\em Deep Subspace clustering Network},
  Advances in Neural Information Processing Systems (2017).

\bibitem{tian2014learning}
F.~Tian, B.~Gao, Q.~Cui, E.~Chen, and T.~Liu,
  {\em Learning deep representations for graph clustering.},
  AAAI (2014),
  pp.~1293--1299.
  
\bibitem{xie2016unsupervised}
J.~Xie, R.~Girshick, and A.~Farhadi,
  {\em Unsupervised deep embedding for clustering analysis},
  International Conference on Machine Learning (2016),
  pp.~478--487.

\bibitem{guo2017deep}
X.~Guo, X.~Liu, E.~Zhu, and J.~Yin,
  {\em Deep Clustering with Convolutional Autoencoders},
  International Conference on Neural Information Processing (2017),
  pp.~373--382.
  
\bibitem{hu2017learning}
 W.~Hu, T.~Miyato, S.~Tokui, E.~Matsumoto, and M.~Sugiyama,
  {\em Learning Discrete Representations via Information Maximizing Self Augmented Training},
  arXiv preprint arXiv:1702.08720 (2017).
  
\bibitem{shaham2018spectralnet}
U.~Shaham, K.~Stanton, H.~Li, R.~Basri, B.~Nadler and Y.~Kluger,
{\em SpectralNet: Spectral Clustering using Deep Neural Networks},
International Conference on Learning Representations (2018).


\bibitem{wilson2016deep}
A.~G. Wilson,  Z.~Hu, R.~Salakhutdinov and E.~P. Xing,
  {\em Deep kernel learning},
  Artificial Intelligence and Statistics (2016),
  pp.~370--378.

\bibitem{wilson2016stochastic}
A.~G. Wilson,  Z.~Hu, R.~Salakhutdinov and E.~P. Xing,
  {\em Stochastic variational deep kernel learning},
  Advances in Neural Information Processing Systems (2016),
  pp.~2586--2594.

\bibitem{wilson2011gaussian}
A.~G. Wilson,  D.~A. Knowles, and Z.~Ghahramani
  {\em Gaussian process regression networks},
  arXiv preprint arXiv:1110.4411 (2011).
  
\bibitem{zhou2004learning}
D.~Zhou, O.~Bousquet, T.~N. Lal, J.~Weston, B.~and Sch{\"o}lkopf,
  {\em Learning with local and global consistency},
  Advances in neural information processing systems (2004),
  pp.~321--328.
  
\bibitem{niu2011dimensionality}
D.~Niu, J.~Dy, and M.~Jordan,
  {\em Dimensionality reduction for spectral clustering},
  Proceedings of the Fourteenth International Conference on Artificial Intelligence and Statistics (2011),
  pp.552--560.
  
\bibitem{wu2018iterative}
C.~Wu, S.~Ioannidis, M.~Sznaier, X.~Li, Xiangyu,D.~Kaeli, David and J.~Dy,
  {\em Iterative Spectral Method for Alternative Clustering},
  International Conference on Artificial Intelligence and Statistics(2018),
  pp.~115--123.

\bibitem{fowlkes2004spectral}
C.~Fowlkes, S.~Belongie, F.~Chung, and J.~Malik,
  {\em Spectral grouping using the Nystrom method},
  IEEE transactions on pattern analysis and machine intelligence (2004),
  pp.~214--225.
  
\bibitem{song2007supervised}
L.~Song, A.~Smola, A.~Gretton K.~M. Borgwardt, and J.~Bedo,
  {\em Supervised feature selection via dependence estimation},
  Proceedings of the 24th international conference on Machine learning (2007),
  pp.~823--830.
\bibitem{vladymyrov2016variational}
M.~Vladymyrov, and M.~Carreira-Perpi{\~n}{\'a}n,
  {\em The Variational Nystrom method for large-scale spectral problems},
  International Conference on Machine Learning (2016),
  pp.~211--220.
  
\bibitem{AbsMahSep2008} 
 P.~A. Absil, R.~Mahony and R.~Sepulchre,
 {\em Optimization Algorithms on Matrix Manifolds},
 Princeton University Press,Princeton, NJ ,(2008)

\bibitem{wen2013feasible}
  Z.~Wen and W.~Yin,
  {\em A feasible method for optimization with orthogonality constraints},
  Mathematical Programming (2013),
  pp.~397--434.
  
\bibitem{horn1990matrix}
R.~A. Horn, and R.~A. Horn, and C.~R. Johnson
  {\em Matrix analysis},
  Cambridge university press (1990).
  
\bibitem{breastcancer}
W.~H. Wolberg,
  {\em Wisconsin breast cancer dataset},
  University of Wisconsin Hospitals (1992).

\bibitem{mangasarian1990cancer}
  O.~L. Mangasarian,
  {\em Cancer diagnosis via linear programming},
  SIAM news (1990),
  pp.18.
  
\bibitem{Dua:2017}
D.~Dheeru, and E.~Karra Taniskidou,
year = "2017",
{\em UCI Machine Learning Repository},
http://archive.ics.uci.edu/ml,
University of California, Irvine, School of Information and Computer Sciences.


\bibitem{bay2000uci}
S.~D. Bay, D.~Kibler, M.~J. Pazzani, and P.~Smyth,
  {\em The UCI KDD archive of large data sets for data mining research and experimentation},
  ACM SIGKDD Explorations Newsletter (2000),
  pp.~81--85.

\bibitem{ding2015yinyang}
Y.~Ding, Y.~Zhao, X.~Shen, M.~Musuvathi, and T.~Mytkowicz,
  {\em Yinyang k-means: A drop-in replacement of the classic k-means with consistent speedup},
  International Conference on Machine Learning (2015),
  pp.~579--587.
  
\bibitem{ng2002spectral}
  A.~Y. Ng, M.~I. Jordan, and Y.~Weiss,
  {\em On spectral clustering: Analysis and an algorithm},
  Advances in Neural Information Processing Systems (2002),
  pp.~849--856.
  
\bibitem{strehl2002cluster}
A.~Strehl and J.~Ghosh,  
{\em Cluster ensembles---a knowledge reuse framework for combining multiple partitions},
Journal of machine learning research (2002),
  pp.~583--617.

\bibitem{dang2010generation}
X.~H. Dang and J.~Bailey,
  {\em Generation of alternative clusterings using the cami approach},
  Proceedings of the 2010 SIAM International Conference on Data Mining (2010),
  pp.~118--129.
  
\bibitem{ross2013nonparametric}
J.~Ross and J.~Dy,
  {\em Nonparametric mixture of Gaussian processes with constraints},
 International Conference on Machine Learning (2013),
  pp.~1346--1354.
  
\bibitem{li2017mmd}
C.~L. Li, W.~C. Chang, Y.~Cheng, Y.~Yang, and B.~P{\'o}czos,
  {\em {MMD GAN}: Towards deeper understanding of moment matching network},
  Advances in Neural Information Processing Systems (2017),
  pp.~2203--2213.

\bibitem{rahimi2008random}
A.~Rahimi and B.~Recht, {\em Random features for large-scale kernel machines}, Advances in Neural Information Processing Systems (2008), pp.~1177--1184.

\bibitem{musco2017recursive}
C.~Musco and C.~Musco, {\em
  Recursive Sampling for the Nystrom Method},
  Advances in Neural Information Processing Systems (2017), pp.~3836--3848.
\end{thebibliography}
\end{document}